\documentclass{article}

\usepackage{arxiv}

\usepackage[utf8]{inputenc} 
\usepackage[T1]{fontenc}    
\usepackage{hyperref}       
\usepackage{url}            
\usepackage{booktabs}       
\usepackage{amsfonts, amsmath}       
\usepackage{amssymb}        
\usepackage{nicefrac}       
\usepackage{microtype}      
\usepackage{lipsum}         
\usepackage{graphicx}
\usepackage[numbers]{natbib}
\usepackage{doi}
\usepackage{enumitem}
\usepackage[ruled,vlined,algosection]{algorithm2e}
\SetAlgoNoLine 
\SetAlgoNoEnd 
\DontPrintSemicolon 
\usepackage{algpseudocode}
\usepackage{float}
\usepackage{caption}
\usepackage{subcaption}
\usepackage{bm}             

\title{Iterative Encoding-Decoding VAEs Anomaly Detection in NOAA's DART Time Series: A Machine Learning Approach for Enhancing Data Integrity for NASA's GRACE-FO Verification and Validation}

\author{ \href{https://orcid.org/0009-0004-0388-9260}{\includegraphics[scale=0.06]{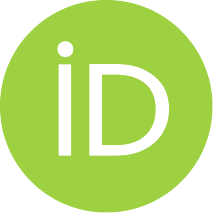}\hspace{1mm}Kevin Lee}\\
    \\
    University of California at Los Angeles and\\
    NASA Jet Propulsion Laboratory
}

\renewcommand{\shorttitle}{Iterative Encoding-Decoding VAEs Anomaly Detection in DART for NASA's GRACE-FO V\&V}

\hypersetup{
pdftitle={Anomaly Detection in NOAA's DART Data Using Iterative Encoding-Decoding VAEs for NASA's GRACE-FO V\&V},
pdfsubject={cs.LG},
pdfauthor={Kevin Lee},
pdfkeywords={Variational Autoencoders, Machine Learning, Deep Learning, Time Series Analysis, Anomaly Detection, DART, GRACE-FO, Tsunami Detection},
}

\begin{document}
\maketitle

\begin{abstract}
NOAA's Deep-ocean Assessment and Reporting of Tsunamis (DART) data are critical for NASA-JPL's tsunami detection, real-time operations, and oceanographic research. However, these time-series data often contain spikes, steps, and drifts that degrade data quality and obscure essential oceanographic features. To address these anomalies, the work introduces an Iterative Encoding-Decoding Variational Autoencoders (Iterative Encoding-Decoding VAEs) model to improve the quality of DART time series. Unlike traditional filtering and thresholding methods that risk distorting inherent signal characteristics, Iterative Encoding-Decoding VAEs progressively remove anomalies while preserving the data's latent structure. A hybrid thresholding approach further retains genuine oceanographic features near boundaries. Applied to complex DART datasets, this approach yields reconstructions that better maintain key oceanic properties compared to classical statistical techniques, offering improved robustness against spike removal and subtle step changes. The resulting high-quality data supports critical verification and validation efforts for the GRACE-FO mission at NASA-JPL, where accurate surface measurements are essential to modeling Earth's gravitational field and global water dynamics. Ultimately, this data processing method enhances tsunami detection and underpins future climate modeling with improved interpretability and reliability.
\end{abstract}


\section{Introduction}
The Deep-ocean Assessment and Reporting of Tsunamis (DART) system, operated by the National Oceanic and Atmospheric Administration (NOAA), serves as the cornerstone of ocean-based tsunami observation and detection worldwide. DART buoys are outfitted with Bottom Pressure Recorders that will continuously monitor the alteration in ocean bottom pressure, enabling the real-time detection of tsunami waves along with other sea-level disturbances \citep{Gonzalez2005DART,Bernard2006DART,Adams2017DART}. These data are of paramount importance not only for immediate hazard assessment but also for long-term climate and oceanographic research. High-quality data are important to improve the understanding of the dynamics of the oceans and validation of satellite-derived measurements, such as those from the NASA-JPL GRACE-FO mission. 

Despite their criticality, many DART time-series data contain both long-term and short-term anomalies that degrade the quality of the data. Such anomalies include sudden spikes, baseline shifts (steps), and gradual drifts. The spikes can arise from instrumental noise, environmental perturbations, or temporary failures of the sensors, while the steps arise from calibration shifts, sensor drift, or true oceanographic phenomena. Both linear and exponential drifts clutter the baseline, possibly resulting or mimicking true long-term oceanic trends \citep{Bernard2006DART}. These anomalies lead to a degradation in the reliability of tsunami detection and impact the accuracy of downstream applications, including climate modeling, ocean mass balance analysis, and verification and validation (V\&V) analyses of the gravity field measurements of GRACE-FO.

Conventional methods of anomaly detection, such as median filters, wavelet transforms, and thresholding methods of various kinds, do not work very well for the variability and complexity of oceanographic time series \citep{Hampel1974InfluenceFunctions,Donoho1995WaveletShrinkage,Nair2011Thresholding}. These methods tend to be over-smooth, resulting in important information lost or misclassify actual oceanographic signals as noise. New methods are to be sought that capture the complexity in the DART data but retain the important character of the signals. 

Recent advances in unsupervised deep learning have achieved considerable success, especially for Variational Autoencoders, in effectively learning complex data distributions and finding anomalies based on reconstruction metrics \citep{An2015VariationalAnomaly,Park2018MultimodalAnomaly}. However, it is uncertain that a single VAEs reconstruction could wash out persistent or hidden anomalies while retaining the original data's complex structure. In light of the above observation, the present paper proposes an improved model called \textit{Iterative Encoding-Decoding Variational Autoencoders (Iterative Encoding-Decoding VAEs)} to solve this signal processing challenge. The Iterative Encoding-Decoding VAEs further refines data representation by iteratively encoding and decoding it, improving the reconstruction accuracy with a reduction in anomalies, hence making the DART time series intact.

In this work, we apply the Iterative Encoding-Decoding VAEs methodology on a very challenging DART dataset, namely, Station 23461 located in the Pacific Ocean, recognized to have several challenging patterns and significant anomalies. Although it has been shown that the algorithmic denoise was effective in the case of Station 23461 presented here, it also behaves well for other DART stations. Because of its nature, the outlined approach presupposes multiple rounds of enhancement with thoroughly worked-out regularization techniques, which can help to distinguish real ocean variability from the artifacts caused by instruments. Improvement in the integrity of data is crucial not only for accurate tsunami detection but also for consistency of the oceanographic measurement with independent satellite observation; such progress helps in the validation of data from the GRACE-FO mission and other integrated Earth system models. 

\textbf{The algorithmic approach contributes to the following fields for satellite applications and Earth science research:}

\textbf{Introduction of Iterative Encoding-Decoding VAEs for anomaly detection:} We introduce a novel system composed of layers of VAEs, which is able to handle abnormal and shifting problems that may arise in DART time series.

\textbf{Better performance than classic methods:} The proposed Iterative Encoding-Decoding VAEs outperforms the traditional despiking and step detection techniques. The integrity of the data remains, with the most important oceanographic features preserved.

\textbf{Validation on a complex DART dataset:} We use data from DART Station 23461 (2022) to show that the model is robust and works well in dealing with complex patterns and different types of anomalies.

\textbf{GRACE-FO mission data integrity:} The Iterative Encoding-Decoding VAEs gives a much cleaner and more reliable output of the DART tool, which can be used in V\&V missions like GRACE-FO to enhance their integrity and further implications in ocean and climate studies.

Through these improvements, Iterative Encoding-Decoding VAEs pushes anomaly detection methods data refinement further with modern algorithms, taking advantage of the computation superiority of graphics processing units (GPUs). It would help benefit more ocean monitoring and Earth science modeling applications.

\section{Background}

\subsection{Variational Autoencoders (VAEs)}

Variational Autoencoders (VAEs) represent one of the key mergers of neural network architectures with probabilistic modeling. It is designed in a way that input data represents in a continuous latent space from which it can later be reconstructed. Both encoding and decoding are performed in a probabilistic manner; hence, this allows the latent space to be conformed to a standard probabilistic distribution-usually Gaussian. Unlike in the case of classic autoencoders, VAEs incorporate a probabilistic framework that lets them generalize even better and further allows generating new samples from the distribution they learned. Overall, VAEs do quite well on all kinds of applications, which involve generation and learning of representation.

In particular, VAEs have been very successful in time-series data analysis with respect to the tasks of anomaly detection and data denoising \citep{An2015VariationalAnomaly,Park2018MultimodalAnomaly}. Because the VAEs learn the representations of data concerning latent patterns, normal fluctuations differ from abnormal events, which resemble sudden spikes. Probabilistic representation gives much better results in the application, where nonlinear dependencies are important and difficult to model using standard linear approaches, like the DART system. Forecast precision is enriched when VAEs are applied in deep oceanographic phenomena, hence enabling decision-making processes linked with climate modeling.

\subsection{Mechanism of VAEs and Iterative Encoding-Decoding VAEs}

\subsubsection{Mechanism of Variational Autoencoders}

The VAEs framework is based on two simple building blocks: the encoder and the decoder \citep{Doersch2016Tutorial,Kingma2019Introduction}. The encoder maps the input, given by $\mathbf{x}$, to a lower-dimensional latent space $\mathbf{z}$, while the decoder reconstructs the input using this latent variable, $\hat{\mathbf{x}}$. Mathematically, the dependence of the encoder and the decoder works as:

\textbf{Encoder (Recognition Model)}: Parameterized by $\phi$, the encoder estimates the posterior $q_\phi(\mathbf{z}|\mathbf{x})$, encoding the data in the latent space to effectively represent input characteristics:
\begin{equation}
q_\phi(\mathbf{z}|\mathbf{x}) = \mathcal{N}\left(\mathbf{z}; \boldsymbol{\mu}_\phi(\mathbf{x}), \boldsymbol{\Sigma}_\phi(\mathbf{x})\right)
\end{equation}
where $\boldsymbol{\mu}_\phi(\mathbf{x})$ and $\boldsymbol{\Sigma}_\phi(\mathbf{x})$ are the mean vector and covariance matrix output by the encoder network parameterized by $\phi$.

\textbf{Decoder (Generative Model)}: Parameterized by $\theta$, the decoder reconstructs the input data through sampling in the latent space, maximizing the likelihood of the observed data:
\begin{equation}
p_\theta(\mathbf{x}|\mathbf{z}) = \mathcal{N}\left(\mathbf{x}; f_\theta(\mathbf{z}), \sigma^2\mathbf{I}\right)
\end{equation}
where $f_\theta(\mathbf{z})$ is the deterministic function modeled by the decoder, and $\sigma^2$ is the variance assumed to be constant.

\textbf{Loss Function}: The approach to setting the loss function combines two terms: one for reconstruction loss to force the model to capture the important input information, and a Kullback-Leibler (KL) divergence term that regularizes the latent space to be close to a predefined prior distribution, usually standard normal:
\begin{equation}
\mathcal{L}(\theta, \phi; \mathbf{x}) = \mathbb{E}_{q_\phi(\mathbf{z}|\mathbf{x})}\left[\log p_\theta(\mathbf{x}|\mathbf{z})\right] - D_{\text{KL}}\left(q_\phi(\mathbf{z}|\mathbf{x}) \| p(\mathbf{z})\right)
\end{equation}
where $D_{\text{KL}}$ denotes the KL divergence, and $p(\mathbf{z}) = \mathcal{N}(\mathbf{0}, \mathbf{I})$ is the prior distribution.

The KL divergence term enforces continuity and smoothness in the latent space, hence coherent reconstructions and the possibility of correct anomaly detection.

\subsubsection{Iterative Encoding-Decoding VAEs and Feature Extraction Enhancement}

The Iterative Encoding-Decoding VAEs is an extension of the basic VAEs architecture \citep{Sohn2015LearningStructured,Bombarelli2018AutomaticChemical}; it utilizes multiple stages of encoding and decoding. The fundamentally useful process of iteration will be particularly enhanced in the latent representation that normally captures features very effectively from complex and noisy data, with the example being the DART time series. The features observed by the model will be refined through several iterations in learning to differentiate normal patterns from anomalous features, with the process realizing optimum spike elimination and assuring maximum performance.

The single-cycle encoding-decoding output is fed back with the Iterative Encoding-Decoding VAEs. In other words, such iterative feedback in the processes allows the latent representation to be developed, thus causing quality improvement within some predefined steps. The iterative approach, within the same essence as the standard VAEs, initiates exploration into the data's underlying structure deeper. Over multiple cycles, the iterative refinement in the latent space allows the model to understand complex patterns and data correlations.

The Iterative Encoding-Decoding VAEs process begins with:
\begin{equation}
\mathbf{z}^{(1)} = q_\phi(\mathbf{z}|\tilde{\mathbf{x}})
\end{equation}
Here, $q_\phi(\mathbf{z}|\tilde{\mathbf{x}})$ represents the encoder network parameterized by $\phi$, which takes the normalized input data $\tilde{\mathbf{x}}$ and maps it to a latent space representation $\mathbf{z}^{(1)}$. The first encoding layer captures the major features of the time series.

The reconstruction from the latent space is taking effects through the decoder network:
\begin{equation}
\hat{\tilde{\mathbf{x}}}^{(1)} = p_\theta(\tilde{\mathbf{x}}|\mathbf{z}^{(1)})
\end{equation}
The decoder network $p_\theta(\tilde{\mathbf{x}}|\mathbf{z}^{(1)})$, parameterized by $\theta$, reconstructs the data from the latent representation, producing the first iteration reconstruction $\hat{\tilde{\mathbf{x}}}^{(1)}$. The reconstruction aims to preserve the essential structure while potentially removing anomalies.

For subsequent iterations ($k > 1$), the encoder processes the previous reconstruction $\hat{\tilde{\mathbf{x}}}^{(k-1)}$ rather than the original input:
\begin{equation}
\mathbf{z}^{(k)} = q_\phi(\mathbf{z}|\hat{\tilde{\mathbf{x}}}^{(k-1)})
\end{equation}
The iterative refinement helps to progressively remove anomalies while maintaining the underlying signal structure \citep{Zhou2017AnomalyDetection}. Each iteration allows the model to focus on increasingly subtle aspects of the time series \citep{Kingma2019Introduction}.

\textbf{Final Reconstruction}: After $n$ iterations, the final reconstruction is obtained from the decoder:
\begin{equation}
\hat{\tilde{\mathbf{x}}}^{(n)} = p_\theta(\tilde{\mathbf{x}}|\mathbf{z}^{(n)})
\end{equation}
This final reconstruction $\hat{\tilde{\mathbf{x}}}^{(n)}$ represents cleaned time series data, where the spike and step anomalies have been handled through iterative encoding and decoding \citep{An2015VariationalAnomaly}. Here, the number of iterations $n$ is determined through the process of training the model to balance the trade-off between the removal of anomalies and preserving legitimate signal features. This ensures that each reconstruction builds upon the previous one, leading to a refined output that retains essential characteristics of the time-series.

\subsection{Computational Considerations}

The Iterative Encoding-Decoding VAEs has particular advantages in handling representation manipulations related to complex datasets, and, thus, are suitable for the reconstruction of time-series data. It is hard for traditional methods operating in this area to encapsulate such complexity. In this regard, an iterative framework enriches the Variational-Autoencoders to take into account the variability of noise that can be present in time series datasets, leading to better reconstructions and improving anomaly detection in time series.

Regularization techniques, such as L2 regularization, help a lot in reducing overfitting during training and further achieving good generalization on previously unseen data. A real implementation is necessary in order to make critical comparisons among the main hyperparameters: latent dimension $d$, the total number of iterations $n$, and the learning rate, regarding the best performance results.

\section{Methods}

An in-depth description of the Iterative Encoding-Decoding VAEs methodology is performed on a case study basis, considering data from DART Station 23461, positioned in the Pacific Ocean in the year 2022. All analyses and results are presented for this dataset; however, the methodology developed in this work can be applied to any DART time-series data. The given data at Station 23461 in the year 2022 is chosen for its highly complicated feature, with many sudden step detortions, intense spike formations, and noise structures, all combined. These have conventionally always been a nightmare to handle with traditional statistical methods. Therefore, this dataset is a perfect test for the manifestation of the robustness and efficiency of the Iterative Encoding-Decoding VAEs algorithm. It is in these step-like gradual transitions, together with abrupt changes in the data, that one obtains a comprehensive framework for model validation regarding an ability to distinguish real oceanographic signals from anomalous spikes.

\subsection{Model Architecture and Implementation}

The Iterative Encoding-Decoding VAEs implementation involves the complex architecture, custom-designed for time-series processing, with special consideration given to handling the temporal dependencies inherent in DART datasets. The crux of the architecture relies on an encoder-decoder mechanism and is further empowered with various techniques that ensure robust performance.

The encoder network begins with an input layer with a window size of 48, which is crucial for capturing the temporal context in the data \citep{Hochreiter1997LSTM,Sutskever2014Sequence}. It then uses three consecutive dense layers of sizes 512, 256, and 128 that use batch normalization and ReLU activations, with dropout rates adjusted according to the layer depth dynamically, defined as:
\begin{equation}
p_{\text{dropout}}(l) = \min(0.1 + 0.05 \times l, 0.3)
\end{equation}
where $l$ is the layer index. This adaptive dropout strategy can prevent overfitting without sacrificing model flexibility.

The final output of the encoder is the latent space representation, which is a 16-dimensional feature vector defined through empirical studies. In preparation for this latent space, parallel dense layers are used to predict the mean ($\boldsymbol{\mu}$) and the log-variance ($\log \boldsymbol{\sigma}^2$), so that the reparameterization trick can be used:
\begin{equation}
\mathbf{z} = \boldsymbol{\mu} + \boldsymbol{\sigma} \odot \boldsymbol{\epsilon}, \quad \boldsymbol{\epsilon} \sim \mathcal{N}(\mathbf{0}, \mathbf{I})
\end{equation}
where $\odot$ denotes element-wise multiplication.

The decoder has a mirrored configuration for the encoder, which rebuilds the input data from the latent variables. It uses dense layers of sizes 128, 256, and 512, respectively, followed by a final output layer equal in size to the input window. Additionally, the decoder allows skip connections in the architecture, which are important for maintaining high-frequency components and temporal sequences:
\begin{equation}
\text{output}_l = \text{Dense}_l(\text{input}_l) + \alpha \times \text{input}_l
\end{equation}
where $\alpha$ is a learnable parameter initialized to 0.8.

In addition, a global skip connection is utiltized to directly link the input to the output to ensure baseline stability and prevent drift:
\begin{equation}
\text{final\_output} = \text{decoder\_output} + \beta \times \text{input\_signal}
\end{equation}
During training, the parameter $\beta$ is dynamically adjusted based on the model's reconstruction confidence, calculated as:
\begin{equation}
\beta_t = \beta_0 \times \exp(-\lambda \times \text{confidence}_t)
\end{equation}
where $\beta_0 = 0.8$ and $\lambda = 0.5$.

\subsection{Training Strategy}

The Iterative Encoding-Decoding VAEs model was implemented from scratch to utilize multi-GPUs for accelerated training. While the datasets used in DART were small, multi-GPUs implementations allowed for parallel processing, hence greatly reducing the duration of training. This enables fast experimentation, hence improvement of the model, an important aspect of developing any anomaly detection system.

\subsubsection{Utilization of Multiple GPUs for Training Acceleration}

In order to increase training and computational efficiency, \texttt{MirroredStrategy} has been used through TensorFlow, which allows multiple GPUs to perform distributed training. The methodology has exploited data parallelism to distribute the load for training equally among the available GPUs and hence reduce the overall training time.

This training process scales the work onto all GPUs and does auto-tuning of batch size to optimal performance. The batch size is scaling linearly with respect to the number of GPUs:
\begin{equation}
\text{batch\_size} = \text{base\_batch\_size} \times \text{num\_gpus}
\end{equation}

where the base batch size was set to 128. Scaling the batch size linearly with the number of GPUs ensures that each GPU is utilized effectively, maximizing computational throughput and reducing the overall training time.

The \textit{all-reduce} algorithm was applied for synchronizing the gradients: all aggregate gradients are updated across all GPUs to maintain model updates consistently. This step is consistent with model convergence, ensuring the best performance once training has been divided across multiple GPUs devices.

\textbf{Training Hardware:} One NVIDIA RTX 3090 and two NVIDIA RTX 3080 Ti GPUs are utilized for the model. Thus, the code distributed training work among those GPUs and reduced overall training time. Multi-GPUs training can process a batch in 23--24 ms per step; therefore, it proves that distributed training accelerated the training of the Iterative Encoding-Decoding VAEs when complex models are built.

Besides, this approach accelerates the training process; it conveys useful insights that can be utilized in the multi-GPUs settings of prospective studies, which will be oriented toward training efficiency improvements for similar models and datasets. Efficient computation applies to running iterative experiments, especially when complex models are built for the Iterative Encoding-Decoding VAEs.

\subsubsection{Optimization and Regularization}

The enhancement and stabilization are critical to establish an optimal setting for the Iterative Encoding-Decoding VAEs model, achieved by well-stabilized convergence. Application of L2 regularization introduced a weight decay of $1 \times 10^{-5}$, penalizing extremely large weights against overfitting. Further, gradient clipping was applied-constrained at a maximum norm of 1.0-to avoid an explosion in the gradient and allow for robust updates and convergence.

A custom learning rate schedule was implemented that updates the learning rate based on the current phase of training:

\begin{equation}
\text{lr} = \text{base\_lr} \times 0.5^{\left\lfloor \frac{\text{epoch}}{\text{decay\_steps}} \right\rfloor}
\end{equation}

where the base learning rate ($\text{base\_lr}$) is set to $1 \times 10^{-4}$ and the decay steps are set to 100. This implies that during the training, an exponential decay in the learning rate is adopted to fine-tune the model parameters.

The model uses the Adam optimizer, which is preferred for non-stationary objectives common when working with time-series data \citep{Kingma2015Adam}. It has prepared this optimizer for the set learning rate schedule and developed it using gradient accumulation to ensure consistent training across multiple GPUs.

It provides the highest computational efficiency and robustness for the training of the Iterative Encoding-Decoding VAEs model and addresses the inherent complexities within the DART time series data very effectively. This optimization is achieved without any degradation of the performance or stability in the model's outputs.

\subsection{Algorithm for Anomaly Detection}

The Iterative Encoding-Decoding VAEs represent an effective anomaly detection system to analyze DART time series data. This framework enables the detection of two types of main categories: spikes and level shifts. The proposed algorithm combines VAEs-based reconstruction with statistical validation methods in order to enhance the robustness of the anomaly detection. Hence, this method ensures accurate and reliable detection of the various types of anomalies present in the data.

\subsubsection{Multi-Scale Anomaly Detection}

The algorithm processes data efficiently by generating data analytics across several time dimensions to identify both short-term spikes and longer-term level shifts. This is accomplished through the analysis of data using a range of window sizes:
\begin{equation}
\text{window\_sizes} = \{48, 480\} \text{ samples}
\end{equation}
where 48 samples target short-term spikes and 480 samples capture level shifts.

\subsubsection{Hybrid Detection Strategy}

The hybrid method for anomaly detection combines both reconstruction-based and statistical techniques to identify anomalies within time series data with a certain anomaly score. The reconstruction error developed in the Iterative Encoding-Decoding VAEs contains the information about the deviation from the expected pattern, while the statistical deviation adds the deviation about the standard statistical behavior. This is harmonized in weighting those elements at $\alpha = 0.7$ to have the benefits of both in order to detect even more subtle anomalies.

\subsubsection{Iterative Encoding-Decoding VAEs Algorithm}

The Iterative Encoding-Decoding VAEs is designed to represent time-series data for facilities iteratively to enhance the performance of anomaly detection and data reconstruction. The process will cycle the model through gradual steps that develop its power to distinguish between typical patterns and anomalies within the data.

The key steps of the Iterative Encoding-Decoding VAEs algorithm are as follows:

\begin{algorithm}[H]
\caption{Iterative Encoding-Decoding VAEs Model Training and Inference}
\SetAlgoLined
\KwIn{Normalized input data $\tilde{\mathbf{x}}$, number of iterations $n$, pre-trained VAEs model parameters $(\phi, \theta)$}
\KwOut{Final reconstruction $\hat{\tilde{\mathbf{x}}}^{(n)}$}

\BlankLine
\textbf{Initialization}\;
Set initial reconstruction: $\hat{\tilde{\mathbf{x}}}^{(0)} = \tilde{\mathbf{x}}$\;

\For{$k = 1$ to $n$}{
    \textbf{Encoding Step:}\;
    Obtain latent representation by encoding the input:\;
    $\mathbf{z}^{(k)} \sim q_\phi(\mathbf{z} | \hat{\tilde{\mathbf{x}}}^{(k-1)})$\;
    
    \textbf{Decoding Step:}\;
    Reconstruct the input from the latent representation:\;
    $\hat{\tilde{\mathbf{x}}}^{(k)} = p_\theta(\hat{\tilde{\mathbf{x}}} | \mathbf{z}^{(k)})$\;
    
    \textbf{Latent Space Refinement:}\;
    Update latent variables to refine representation:\;
    $\mathbf{z}^{(k)} = \alpha \mathbf{z}^{(k)} + (1 - \alpha) \mathbf{z}^{(k-1)}$\;
    
    \textbf{Stopping Criterion Check (Optional):}\;
    Evaluate convergence based on reconstruction loss or other metrics\;
    \If{Convergence criterion met}{
        \textbf{break}\;
    }
}

\BlankLine
\Return{$\hat{\tilde{\mathbf{x}}}^{(n)}$}
\end{algorithm}

In the algorithm:

\textbf{Initialization}: The algorithm initializes the normalized input data $\tilde{\mathbf{x}}$ as the first iteration's reconstructed output $\hat{\tilde{\mathbf{x}}}^{(0)}$.

\textbf{Encoding Step}: At each iteration $k$, the model encodes the current reconstruction $\hat{\tilde{\mathbf{x}}}^{(k-1)}$ to obtain an updated latent representation $\mathbf{z}^{(k)}$. This step captures the underlying features of the data, focusing on the refinement of the representation to highlight anomalies.

\textbf{Decoding Step}: The decoding step reconstructs the input data from the updated latent representation, producing $\hat{\tilde{\mathbf{x}}}^{(k)}$. This reconstruction aims to preserve the essential data features while suppressing the anomalies.

\textbf{Latent Space Refinement}: An optional refinement step combines the current and previous latent representations using a weighting factor $\alpha$ (e.g., $\alpha = 0.5$) to smooth the updates and improve stability. This blending helps the model to converge more effectively by considering both the new and prior latent contents in the data.

\textbf{Stopping Criterion}: An optional convergence check can be included to terminate the iterations early if the reconstruction loss or other evaluation metrics indicate that further iterations will not significantly improve the results.

The iterative process gradually improves the model in capturing complex patterns and anomalies present in the data. This, effectively, separates normal data patterns from anomalies by repetition of refinement in latent space and reconstruction, thereby improving anomaly detection performance.

\section{Iterative Encoding-Decoding VAEs Anomaly Detection Process}

All steps involved in the anomaly detection process, from data preparation to the finalization of the process and how it works with the Iterative Encoding-Decoding VAEs algorithm, are presented in this section.

\subsection{Overview of the Process}

The Iterative Encoding-Decoding VAEs Anomaly Detection Process consists of five main steps, \textbf{Data Preprocessing}, \textbf{Iterative Encoding-Decoding VAEs Training}, \textbf{Anomaly Detection}, \textbf{Iterative Refinement}, and \textbf{Post-processing}. Each step is crucial in ensuring accurate anomaly detection and high-quality data reconstruction.

\subsection{Detailed Methodology}

\subsubsection{Step 1: Data Preprocessing}

The process begins with thorough data cleaning to prepare the time-series data, $\mathbf{x}$, for analysis:

\textbf{Removal of Flagged Values}: Any flagged or invalid data points by NOAA's data system (e.g., placeholder values like 9999) are removed from the dataset to eliminate obvious errors.

\textbf{Gap Interpolation}: Missing data points are filled using linear interpolation to maintain temporal continuity, which is essential for data processing in the model.

\textbf{Normalization}: Data are normalized using z-score normalization to ensure consistent scaling and to facilitate the training of the VAEs:
\begin{equation}
\tilde{\mathbf{x}} = \frac{\mathbf{x} - \mu_x}{\sigma_x}
\end{equation}
where $\mu_x$ and $\sigma_x$ are the mean and standard deviation of $\mathbf{x}$, respectively. Normalization helps in stabilizing the learning process and in handling numerical issues during optimization.

\subsubsection{Step 2: Iterative Encoding-Decoding VAEs Training}

At this stage, a VAEs with a specified latent dimension $d$ is initialized and trained to reconstruct the normalized data $\tilde{\mathbf{x}}$. The goal is to learn a compressed representation of the data that captures the underlying patterns while being robust to anomalies.

\textbf{Model Initialization}: The VAEs is initialized with encoder parameters $\phi$ and decoder parameters $\theta$, and a latent dimension $d=16$. The choice of $d$ balances the model's capacity to capture essential features without overfitting to noise.

\textbf{Training Loop}: The model is trained over $e=1000$ epochs using mini-batch gradient descent:

\textbf{i. Encoding}: For each batch, the encoder maps input data to the latent space:
\begin{equation}
\mathbf{z} \sim q_\phi(\mathbf{z}|\tilde{\mathbf{x}})
\end{equation}
where $q_\phi(\mathbf{z}|\tilde{\mathbf{x}})$ is the approximate posterior distribution modeled by the encoder.
    
\textbf{ii. Decoding}: The decoder reconstructs the input from the latent representation:
\begin{equation}
\hat{\tilde{\mathbf{x}}} = p_\theta(\tilde{\mathbf{x}}|\mathbf{z})
\end{equation}
where $p_\theta(\tilde{\mathbf{x}}|\mathbf{z})$ is the likelihood modeled by the decoder.
    
\textbf{iii. Loss Computation and Weight Update}: The model parameters are updated using the Adam optimizer to minimize the VAEs loss function, which consists of the reconstruction loss and the Kullback-Leibler (KL) divergence regularization term:
\begin{equation}
\mathcal{L}(\theta, \phi; \tilde{\mathbf{x}}) = \mathbb{E}_{q_\phi(\mathbf{z}|\tilde{\mathbf{x}})}\left[ -\log p_\theta(\tilde{\mathbf{x}}|\mathbf{z}) \right] + D_{\text{KL}}\left( q_\phi(\mathbf{z}|\tilde{\mathbf{x}}) \parallel p(\mathbf{z}) \right)
\end{equation}
where $p(\mathbf{z})$ is the prior distribution over the latent variables.

\subsubsection{Step 3: Anomaly Detection}

After training, the model proceeds to detect anomalies in the data. Anomalies are identified based on discrepancies between the original data and the model's reconstruction; later, we can use statistical methods with low computation requirements to detect significant deviations.

\textbf{Reconstruction Error Computation}: The reconstruction error for each data point is calculated:
\begin{equation}
\text{RE}_i = \left| \tilde{x}_i - \hat{\tilde{x}}_i \right|
\end{equation}
A high reconstruction error indicates that the model is unable to accurately reconstruct the data point, suggesting a potential anomaly.
    
\textbf{Thresholding for Anomaly Detection}: Anomalies are detected by applying a threshold to the reconstruction errors. The threshold can be determined using statistical measures, such as the mean and standard deviation of the reconstruction errors:
\begin{equation}
\tau_{\text{re}} = \mu_{\text{RE}} + \kappa \sigma_{\text{RE}}
\end{equation}
where $\mu_{\text{RE}}$ and $\sigma_{\text{RE}}$ are the mean and standard deviation of the reconstruction errors, and $\kappa$ is a scalar (e.g., $\kappa = 3.0$).

\textbf{Spike Detection Using Statistical Methods}:

\textbf{i.} A moving median and standard deviation are calculated using a window size of $w_s=48$ samples to capture local variations.

\textbf{ii.} Data points that deviate significantly from the rolling median are marked as spikes:
\begin{equation}
\mathbf{M}_s = \left\{ i \, \bigg| \, \left| x_i - \text{median}_i \right| > \tau_s \times \text{std}_i \right\}
\end{equation}
where $\tau_s$ is a threshold parameter (e.g., $\tau_s = 3.0$), and $\text{std}_i$ is the rolling standard deviation at point $i$.
    
\textbf{Step Detection Using Statistical Methods}:

\textbf{i.} Level shifts (steps) are detected using a larger window size of $w_l=480$ samples to capture long-term changes.

\textbf{ii.} Significant changes in the data's baseline are identified by comparing mean values in adjacent windows:
    \begin{equation}
    \Delta \mu_i = \left| \mu_{i-w_l/2}^{i} - \mu_{i}^{i+w_l/2} \right|
    \end{equation}
    If $\Delta \mu_i > \tau_l$, a level shift is detected at position $i$, where $\tau_l$ is a predefined threshold (e.g., $\tau_l = 0.05$).
    
\textbf{iii.} The detected positions are stored in a step mask $\mathbf{M}_l$.

These combined methods through reconstruction errors with the VAEs enhance the detection of both subtle and significant anomalies.

\subsubsection{Step 4: Iterative Refinement}

The Iterative Encoding-Decoding VAEs algorithm refines the data representation over multiple iterations, improving the model's ability to reconstruct the data and detect anomalies with high accuracy.

\textbf{Initialization}: Set initial reconstruction $\hat{\tilde{\mathbf{x}}}^{(0)} = \tilde{\mathbf{x}}$.

\textbf{Iterative Process}: For each iteration $k$ from 1 to $n=10$:

\textbf{i. Encoding Step}: Encode the current reconstruction:
\begin{equation}
\mathbf{z}^{(k)} \sim q_\phi(\mathbf{z}|\hat{\tilde{\mathbf{x}}}^{(k-1)})
\end{equation}

\textbf{ii. Decoding Step}: Decode to obtain a new reconstruction:
\begin{equation}
\hat{\tilde{\mathbf{x}}}^{(k)} = p_\theta(\hat{\tilde{\mathbf{x}}}^{(k-1)}|\mathbf{z}^{(k)})
\end{equation}

\textbf{iii. Anomaly Mask Update}:

Compute the reconstruction error at iteration $k$:
    \begin{equation}
    \text{RE}_i^{(k)} = \left| \hat{\tilde{x}}_i^{(k)} - \hat{\tilde{x}}_i^{(k-1)} \right|
    \end{equation}

Update the anomaly masks $\mathbf{M}_s$ and $\mathbf{M}_l$ by comparing $\text{RE}_i^{(k)}$ to dynamic thresholds that may decrease over iterations to focus on finer anomalies.
    
\textbf{iv. Data Correction}:

For data points identified as anomalies, replace them with the reconstructed values:
    \begin{equation}
    \hat{\tilde{x}}_i^{(k)} = \begin{cases}
    \hat{\tilde{x}}_i^{(k)}, & \text{if } i \in \mathbf{M}_s^{(k)} \cup \mathbf{M}_l^{(k)} \\
    \hat{\tilde{x}}_i^{(k-1)}, & \text{otherwise}
    \end{cases}
    \end{equation}

This step progressively refines the reconstruction by correcting anomalies while preserving normal data points.

The iterative refinement allows the model to increasingly better reconstruct its data insofar as this is done to reduce the influence of anomalies in subsequent iterations. For each subsequent iteration, updating anomaly masks and correcting data enforces the model to be more and more sensitive (even to very fine anomalies that were missed at the very beginning).

\subsubsection{Step 5: Post-processing}

After several iterations of refining and rectifying anomalies, it can now proceed with ensuring that the derived time series is clean as well as smooth, stable, and ready for use. Post-processing techniques help in converting the reconstructed data to such a form where the essential oceanographic features are retained while the artifacts, which are introduced during iterations, are removed.

\textbf{Smoothing}: The reconstructed data are subjected to Gaussian smoothing filter with a window size of 6. This step of smoothing is done to eliminate minor, transient fluctuations (after iterative refinements) that may arise in general. Such minor fluctuations removed by smoothing actually impede in perceiving the continuity in time series behavior for identifying genuine oceanographic signals like tides or long-term trends.

\textbf{Validation of Detected Steps}: The Iterative Encoding-Decoding VAEs and statistical approaches corrected most step-like anomalies. Some of the shifts that were identified may be considered borderline cases or, on the other hand, data points affected by overlapping anomalies. Threshold adjustment or excluding particular periods would ensure that while real oceanographic changes are preserved, false positives are minimized.

\textbf{Anomaly Merging}: Such merged disjoint anomaly segments may emanate from the detection of multiple anomalies at times when several anomalies are located nearby. Consequently, the next step after anomaly detection is to merge them into single continuous segments to facilitate irregularity representation. Merging in this manner not only tidies the time series but also makes it easier to comprehend the whole pattern of anomalies present in the dataset.

\textbf{Denormalization}: Here, the processed signal should be denormalized to bring back the original scale. This is again essential for practical use:
\begin{equation}
\hat{\mathbf{x}} = \hat{\tilde{\mathbf{x}}}^{(n)} \sigma_x + \mu_x
\end{equation}

Altogether, these post-processing steps shape the output of the Iterative Encoding-Decoding VAEs pipeline would ensure the time-series data is then free from anomalous spikes, steps, or drift and simultaneously clean.

\subsection{Algorithm Summary}

The complete Iterative Encoding-Decoding VAEs Anomaly Detection Process is summarized in Algorithm \ref{alg:iterative_VAEs_anomaly_detection}.

\begin{algorithm}[H]
\caption{Iterative Encoding-Decoding VAEs Anomaly Detection Process}
\label{alg:iterative_VAEs_anomaly_detection}
\SetAlgoLined
\KwIn{Time series data $\mathbf{x}$, latent dimension $d=16$, iterations $n=10$, epochs $e=1000$}
\KwOut{Cleaned data $\hat{\mathbf{x}}$, spike mask $\mathbf{M}_s$, step mask $\mathbf{M}_l$}

\BlankLine
\textbf{Step 1: Data Preprocessing}\;
\Begin{
    Remove flagged values (e.g., 9999)\;
    Interpolate gaps using linear interpolation\;
    Normalize: $\tilde{\mathbf{x}} = (\mathbf{x} - \mu_x) / \sigma_x$\;
}

\BlankLine
\textbf{Step 2: Iterative Encoding-Decoding VAEs Training}\;
\Begin{
    Initialize VAEs with parameters $(\phi, \theta)$ and latent dimension $d$\;
    \For{epoch $=$ 1 to $e$}{
        \For{each batch in $\tilde{\mathbf{x}}$}{
            Encode: $\mathbf{z} \sim q_\phi(\mathbf{z}|\tilde{\mathbf{x}}_{\text{batch}})$\;
            Decode: $\hat{\tilde{\mathbf{x}}}_{\text{batch}} = p_\theta(\tilde{\mathbf{x}}_{\text{batch}}|\mathbf{z})$\;
            Compute loss $\mathcal{L}$ and update weights using Adam optimizer\;
        }
    }
}

\BlankLine
\textbf{Step 3: Anomaly Detection}\;
\Begin{
    Compute reconstruction errors: $\text{RE}_i = \left| \tilde{x}_i - \hat{\tilde{x}}_i \right|$\;
    Set anomaly threshold: $\tau_{\text{re}} = \mu_{\text{RE}} + \kappa \sigma_{\text{RE}}$\;
    Identify initial anomalies: $\mathbf{A} = \{ i \, | \, \text{RE}_i > \tau_{\text{re}} \}$\;
    Detect spikes using statistical thresholding with window size $w_s$\;
    Detect steps using mean shifts with window size $w_l$\;
    Initialize anomaly masks $\mathbf{M}_s$ and $\mathbf{M}_l$\;
}

\BlankLine
\textbf{Step 4: Iterative Refinement}\;
\Begin{
    Set initial reconstruction: $\hat{\tilde{\mathbf{x}}}^{(0)} = \tilde{\mathbf{x}}$\;
    \For{$k = 1$ to $n$}{
        Encode: $\mathbf{z}^{(k)} \sim q_\phi(\mathbf{z}|\hat{\tilde{\mathbf{x}}}^{(k-1)})$\;
        Decode: $\hat{\tilde{\mathbf{x}}}^{(k)} = p_\theta(\hat{\tilde{\mathbf{x}}}^{(k-1)}|\mathbf{z}^{(k)})$\;
        Compute reconstruction errors: $\text{RE}^{(k)}$\;
        Update anomaly thresholds and masks $\mathbf{M}_s^{(k)}$, $\mathbf{M}_l^{(k)}$\;
        Correct anomalies in $\hat{\tilde{\mathbf{x}}}^{(k)}$\;
    }
}

\BlankLine
\textbf{Step 5: Post-processing}\;
\Begin{
    Apply Gaussian smoothing to $\hat{\tilde{\mathbf{x}}}^{(n)}$\;
    Validate and adjust detected steps using temporal context\;
    Merge overlapping anomalies in $\mathbf{M}_s$ and $\mathbf{M}_l$\;
    Denormalize: $\hat{\mathbf{x}} = \hat{\tilde{\mathbf{x}}}^{(n)} \sigma_x + \mu_x$\;
}

\Return{$\hat{\mathbf{x}}$, $\mathbf{M}_s$, $\mathbf{M}_l$}
\end{algorithm}

\subsection{Discussion of the Process}

The anomaly detection process combines the deep learning algorithm, VAEs, with statistical approaches for accurate identification and correction of anomalies in time-series data. This process utilizes a balance between accurately pinning down anomalies and sufficiently reconstructing data. It is appropriate for applications that demand precise analyses of time-series data.

\textbf{i. Conventional Statistical and Machine Learning Methods Integration}: The utilization of moving statistics for initial anomaly detection complements well the ability of VAEs to model complex data distributions for the robustness of the detection process.

\textbf{ii. Iterative Refinement Mechanism}: The model grasps the ability to progressively make reconstructions better and the detection of anomalies, finally making all anomalies—apparently those not directly reconciled in the initial processing—reachable.

\textbf{iii. Adaptive Anomaly Thresholding:} Updating anomaly masks and thresholds over iterations based on reconstruction errors. The process automatically adapts to data and thereby enhances detection accuracy.

\textbf{iv. Preservation of Data Integrity:} It maintains the salient characteristics of the original data and avoids over-smoothing or distortion.

\textbf{v. Scalability and Efficiency}: It is implemented to be computationally efficient for scalability to large datasets while utilizing GPU acceleration for faster training and inference.

\subsection{Data Preprocessing and Memory Management}

Efficient data preprocessing and memory management are very essential when dealing with DART time series data at scale. This will ensure that the Iterative Encoding-Decoding VAEs model runs well within the limitations of the hardware resources at hand.

\subsubsection{Data Preprocessing Pipeline}

Several major steps are introduced in the data preprocessing pipeline for an effective preparation of data for both model training and anomaly detection processes. These steps include:

\textbf{Normalization}: Data are standardized using z-score normalization to ensure consistent scaling:
\begin{equation}
x_{\text{norm}} = \frac{x - \mu}{\sigma}
\end{equation}
where $\mu$ and $\sigma$ are computed using robust statistics to minimize the impact of outliers.

\textbf{Sliding Window}: Implementation of overlapping windows with size $w=48$ and stride $s=1$ to capture temporal dependencies:
\begin{equation}
X_{\text{window}}^i = [x_i, x_{i+1}, \ldots, x_{i+w-1}]
\end{equation}

\textbf{Missing Value Handling}: Gaps in the time series are addressed using forward-fill followed by backward-fill interpolation to maintain temporal continuity.

\subsubsection{GPU Memory Optimization}

To maximize the efficiency of GPU resource utilization, several advanced memory management strategies are implemented:

\textbf{Gradient Accumulation}: This technique involves accumulating gradients over $n$ mini-batches before applying updates, effectively reducing memory usage per batch:
\begin{equation}
\mathbf{g}_t = \frac{1}{n}\sum_{i=1}^n \mathbf{g}_i
\end{equation}
where $\mathbf{g}_i$ represents the gradients of individual batches.

\textbf{Memory-Efficient Backpropagation}: Gradient checkpointing is employed to balance computation and memory usage. This approach reduces the peak memory requirement from $O(N)$ to $O(\sqrt{N})$, where $N$ is the number of layers:
\begin{equation}
\text{Memory}_{\text{peak}} = O(\sqrt{N}) \quad \text{instead of} \quad O(N)
\end{equation}

The gradient checkpointing strategically saves intermediate activations only at certain layers, recalculating them during the backward pass. This balances computation and memory and allows training of the models without exceeding memory limits.

\subsection{Model Refinement and Parameter Tuning}

The Iterative Encoding-Decoding VAEs model hyperparameter and structural decision space have to be explored with a wide range of experiments to find the best setting for optimizing performance on DART time series datasets.

\subsubsection{Training Parameters}

The Iterative Encoding-Decoding VAEs model was trained with optimizations for achieving efficiency in convergence and stable performances. The following key parameters were fine-tuned:

\textbf{Learning Rate Schedule}: A dynamic learning rate schedule was employed to achieve gradual convergence by each epoch. The learning rate at each epoch is defined by:
\begin{equation}
\text{lr}_{\text{epoch}} = 10^{-4} \times 0.5^{\left\lfloor \frac{\text{epoch}}{100} \right\rfloor}
\end{equation}
With this exponential decay strategy, gradually decreasing the learning rate promotes stability and precision in later training stages.

\textbf{Batch Size Optimization}: The batch size was dynamically adjusted to maximize GPU utilization while respecting memory constraints:
\begin{equation}
\text{batch\_size} = \min(128 \times N_{\text{GPUs}}, \text{available\_memory})
\end{equation}
This approach ensures that the training process is both efficient and scalable across different hardware configurations.

\textbf{Iteration Count}: Actual values for training iterations and epochs were determined through the training process for model refinement. The training epochs had been increased from an initial 100 to 1000 to allow better model training. Refinement iterations were increased to 10 from an initial 3 for better capturing the model of complex patterns. Also, early stopping patience was set to 30 epochs to avoid overfitting but at the same time guaranteeing sufficient training.

\section{Results and Analysis}

The comprehensive visual analysis of the Iterative Encoding-Decoding VAEs' performance gives valuable insights into its effectiveness when processing DART time-series data. The model proves strong capabilities in anomaly detection and data cleaning as well as pattern recognition, as evidenced by the following analyses.

\subsection{Water Level Analysis}

The water level data for DART Station 23461 in 2022 represents the most extreme case among DART time-series datasets, featuring various abrupt shift steps, distortions, spikes, and gaps in the data throughout the year. This dataset is of a much higher degree of complexity and variability than most other DART datasets, which would tend to be more regular and stable in their behavior. This fact poses special problems in analytical treatment and physical interpretation of the data. Such a mixture of anomalies and irregularities within a dataset results in a perfect situation for assessing the robustness and effectiveness of the data processing techniques.

\begin{figure}[H]
    \centering
    \includegraphics[width=1\textwidth]{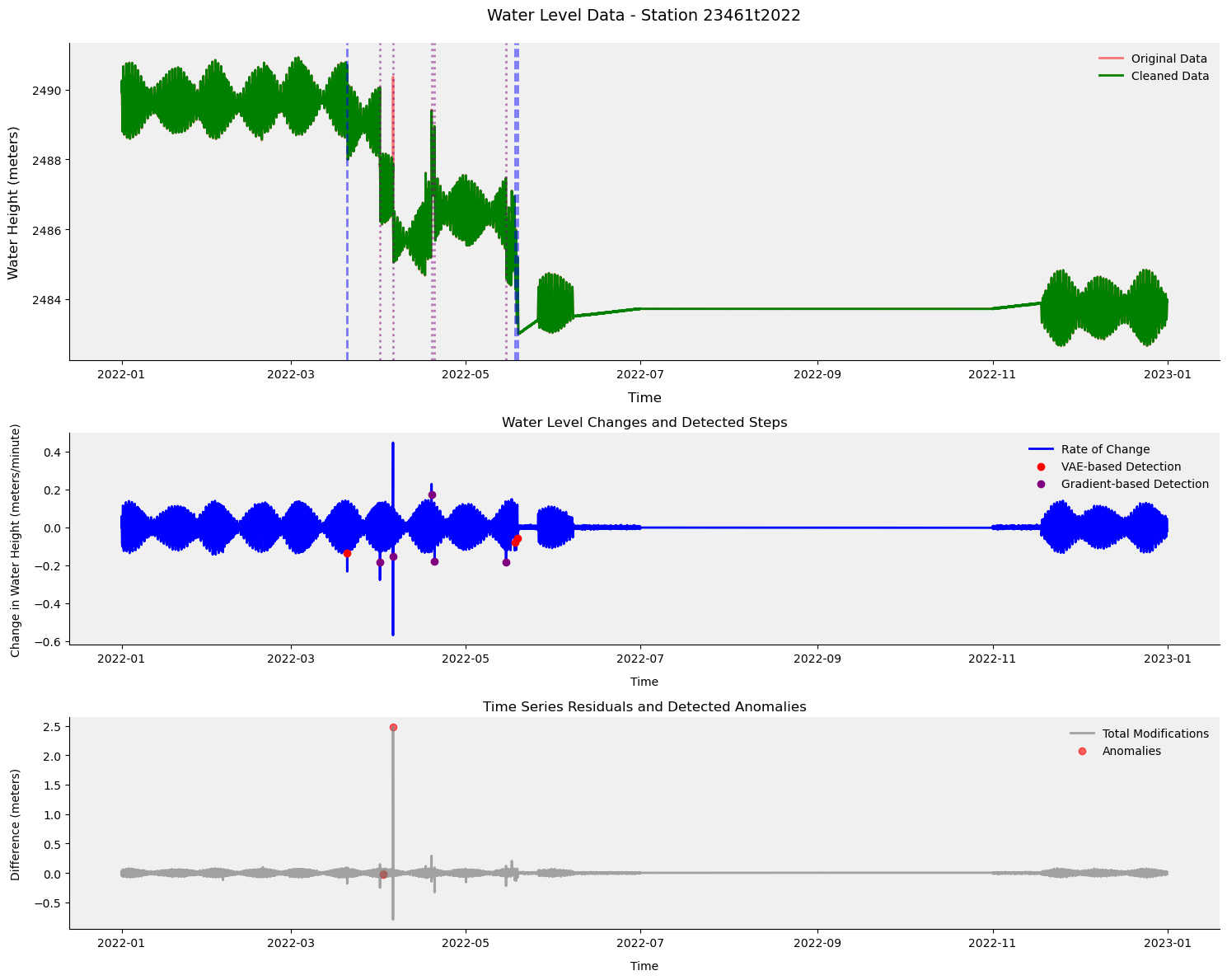} 
    \caption{DART Water Level Data, 23461t2022: Original vs. Cleaned Data}
    \label{fig:water_level}
\end{figure}

The Iterative Encoding-Decoding VAEs model excels in addressing this challenge. It effectively reconstructs the time series, aligning disparate segments back to their original baseline while successfully detaching and removing spikes. Additionally, the use of VAEs and gradient-based algorithms enables robust step detection, resolving this issue within the DART data. This demonstrates the Iterative Encoding-Decoding VAEs's potential in solving complex time series problems, with applicability extending to other domains. The model's proficiency in preserving essential oceanographic signals is evident, maintaining natural fluctuations and safeguarding tidal oscillations, particularly during January–March and November–December 2022. Notably, it accurately identifies abrupt shifts in April–May 2022, distinguishing between genuine level transitions and erroneous spikes, as shown in Figure \ref{fig:water_level}.

\subsection{Rate of Change Analysis}

The rate-of-change analysis shows that the Iterative Encoding-Decoding VAEs model is good at detecting anomalies using a combination of VAEs and gradient-based methods. This reveals the model's ability to recognize fluctuations in water level at $-0.6$ to $0.4$ meters per minute change rate. This precision in detecting rate changes in crucial for distinguishing between normal variability and abnormal spikes in the data.

This ability of the algorithm to distinguish well between inherent fluctuations and abnormal spikes is quite manifestly proved, especially in the critical transition months of April and May. During this time, a clear representation of sudden shifts is provided by the model. This underscores the robustness of Iterative Encoding-Decoding VAEs.

\subsection{Residual Analysis}

\begin{figure}[H]
    \centering
    \includegraphics[width=0.8\textwidth]{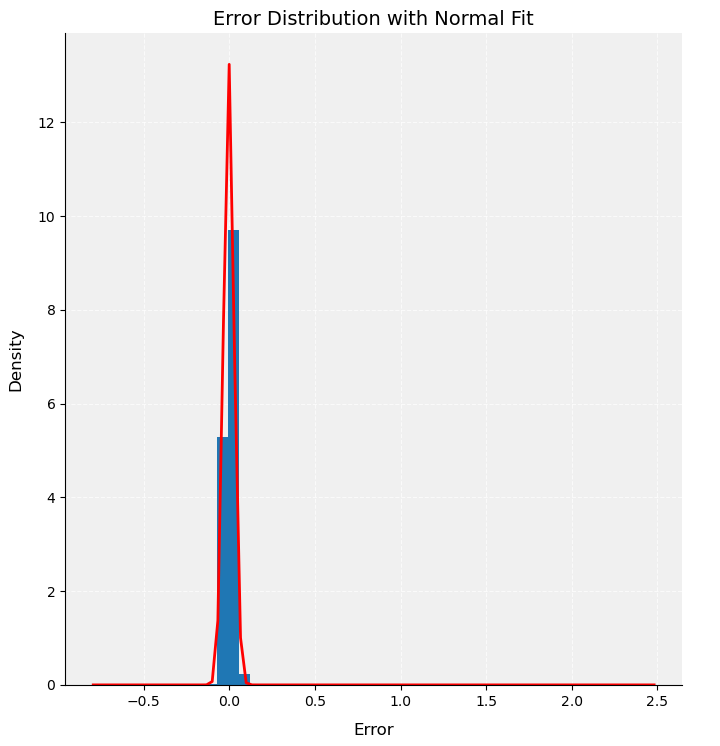}
    \caption{Time Series Residuals and Detected Anomalies}
    \label{fig:residual_analysis}
\end{figure}

The residual plot analysis in Figure \ref{fig:residual_analysis} illustrates how well the Iterative Encoding-Decoding VAEs algorithm can detect spikes and correct anomalies. An extremely clear separation of anomalies from background variations is observed in this plot, with most corrections being well within $\pm 0.5$ meters. This implies that the model is very accurate in aiming at and correcting anomalies, and not overfitting the noise or irrelevant fluctuations.

Anomalies of large magnitudes, up to $2.5$ meters, are identified and properly corrected. The model demonstrated a very strong handling of extreme deviations within the time series. This is important to ensure that the integrity of data is not compromised, especially for datasets characterized by complex and erratic patterns.

The consistent performance in time series for both training and residual analysis in the model implies that Iterative Encoding-Decoding VAEs can generalize well across different segments of the data. Particularly, this is important in use cases that require reliable anomaly detection and correction.

Results from the residual analysis can prove that the Iterative Encoding-Decoding VAEs model is not only great at performances in anomaly detection and correction but also at keeping the primary characteristics of the original data. This balance of anomaly correction and data integrity is evidence of the model's great performance and ability can be extended to other complex time-series datasets.

\subsection{Training Performance}

\begin{figure}[H]
    \centering
    \includegraphics[width=1\textwidth]{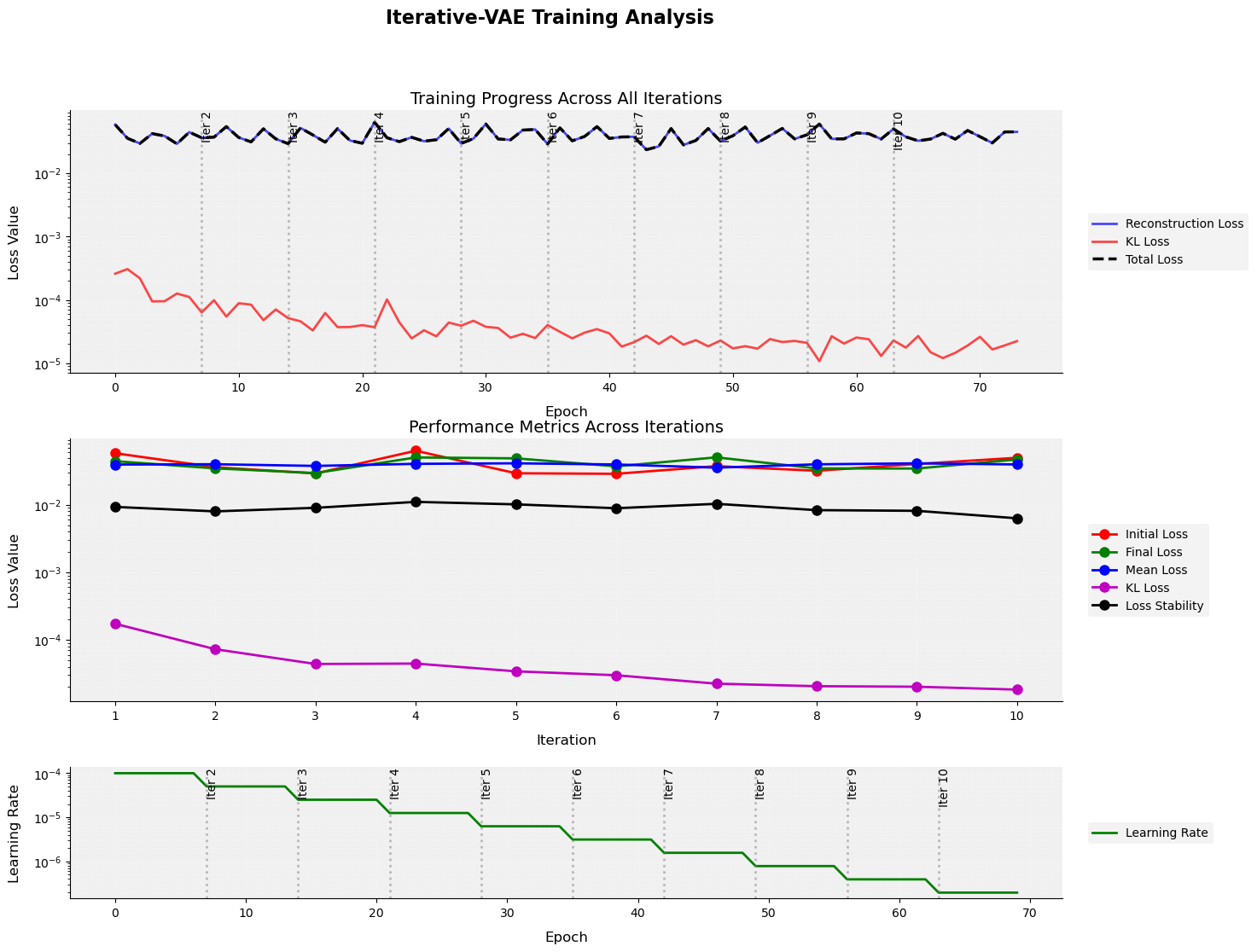}
    \caption{Training History}
    \label{fig:training_performance}
\end{figure}

The training history plot, shown in Figure \ref{fig:training_performance}, provides valuable insights into the Iterative Encoding-Decoding VAEs's training dynamics. The plot reveals a steady convergence, characterized by consistent improvements in loss metrics throughout 10 iterations. This pattern indicates several key aspects of the model's training process.

The gradual reduction in loss values implies that the model is learning the patterns in the data well, without major fluctuations or instability. This stability is crucial for ensuring that the model does not overfit the noise or irrelevant features while keeping its generalization capabilities.

Moreover, given that loss metrics decline smoothly, it implies a properly balanced learning rate schedule. This enables the model to make updates to parameters while still, meaningfully reducing loss without it veering away from an optimum solution. This balance is essential to ensure that the model converges towards a robust solution. The consistency of improvement across all iterations shows the resilience of the Iterative Encoding-Decoding VAEs methodology.

Finally, the iterative training process wherein multiple cycles of encoding and decoding take place is validated to have convergence. This allows the model to gradually improve its feature extraction capabilities, which amounts to accurate anomaly detection and data reconstruction. The history plot during training underscores the robust model training and validation attainability of the Iterative Encoding-Decoding VAEs.

\subsection{Latent Space Visualization}

\begin{figure}[H]
    \centering
    \includegraphics[width=0.8\textwidth]{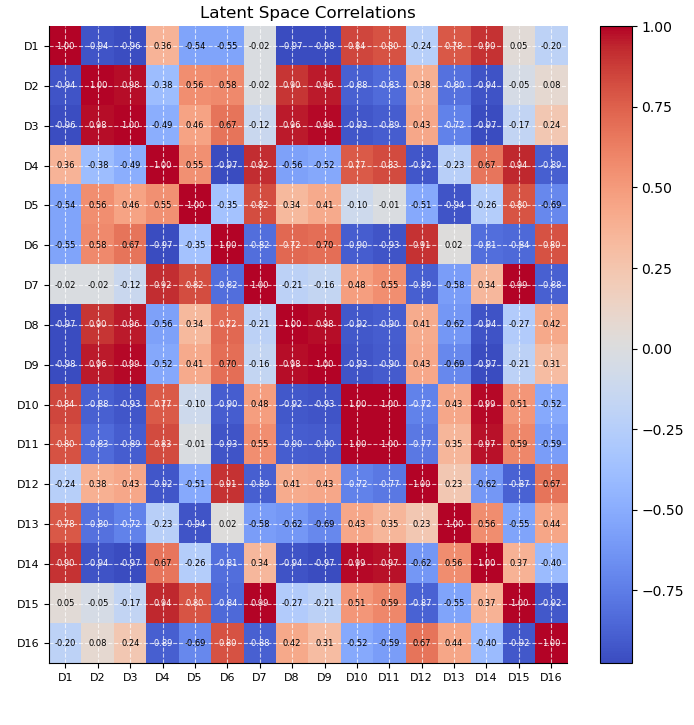}
    \caption{Latent Space Visualization: Clustering of Normal vs. Anomalous Patterns}
    \label{fig:latent_space}
\end{figure}

Figure \ref{fig:latent_space} shows the latent space visualization, offering a detailed view of how the model organizes and interprets DART time-series data. The well-defined clustering of data points in the latent space reflects the model's ability to discern between normal and anomalous patterns. Normal points are grouped within distinct clusters, while anomalous points lie at a distance from the normal points, which shows their deviation from what is considered normal. In this regard, the clear gap separates an effective anomaly detection, showing it can reliably identify outliers and irregularities within the dataset.

Moreover, the smooth transitions within clusters that are observed indicate that continuity and coherence are maintained by the model in its latent representations. This is crucial to ensure that the model can learn the relations of order by preserving the chronological relationships between data points. Such coherence is very important for exact data reconstruction and enhances the ability of the model to detect subtle anomalies.

The latent space visualization proves that the model can adequately handle complicated and noisy datasets by very effectively separating noise elements from genuine signals; the Iterative Encoding-Decoding VAEs guarantees that its representations are both meaningful and informative. This becomes particularly germane in applications that place much demand on data integrity.

On the whole, the latent space visualization proves Iterative Encoding-Decoding VAEs's concept and approach, showing in actuality the potential to manage very complex time-series data with a high level of accuracy and reliability. The ability of the model to learn coherent and separate latent representations further asserts its relevance in many domains for which strong anomaly detection and integrity of data are prime requirements.

\section{Challenges and Solutions in Implementing Iterative Variational Autoencoders}

When implementing the Iterative Encoding-Decoding VAEs to process DART time-series data, several peculiar challenges arose and needed to be coped with. These included efficient handling of large datasets, training stabilization, and keeping the data integrity in the process of modeling. In such an application as the detection of tsunamis and oceanographic surveillance, the assured and exact despiked data come to the fore as especially significant. This section brings forward major technical obstacles that came in the way of developing Iterative Encoding-Decoding VAEs along with the innovative solutions contrived to surmount those. Overcoming these challenges would enable the Iterative Encoding-Decoding VAEs model to have robust performance with high-quality data reconstruction, vital in practice for real-world implementations.

\subsection{GPU Memory Management}

During processing the DART time-series data, the effective management of GPU memory is useful for handling the datasets. This section discusses the strategies employed in improving memory utilization and ensuring steady performance of training.

\subsubsection{Memory-Efficient Data Loading}

The dynamic streaming data pipeline was built using TensorFlow's \texttt{tf.data} API to manage large datasets gracefully with the available RAM capacity. The approach adopted in this work involved the use of memory mapping and prefetch optimization to actually be able to facilitate data handling:

\textbf{Streaming Data Pipeline}: Implemented with a buffer size dynamically set to the lesser of 50,000 or the dataset size, ensuring efficient data shuffling and loading.

\textbf{Prefetch Optimization}: Utilized \texttt{dataset.prefetch(tf.data.AUTOTUNE)} to overlap data preprocessing and model execution, thereby reducing idle time and improving throughput.

\textbf{Memory Mapping}: For datasets larger than available RAM, data is processed in chunks, with the chunk size calculated as:
\begin{equation}
\text{chunk\_size} = \frac{\text{available\_memory}}{3 \times \text{num\_gpus}}
\end{equation}
This ensures that data is loaded efficiently without exceeding memory limits.

\subsubsection{Batch Size Optimization}

To optimize GPU utilization, we use dynamic batch size adjustment for training. This adjustment would allow dynamic adjustment of batch sizes so that GPU usage is maximized in such a way that memory is not exceeded. Dynamic batch sizing helps scale processing more data in parallel for better performance and resource utilization:

\textbf{Initial Estimation}: The maximum batch size is estimated based on available memory and a safety factor:
\begin{equation}
B_{\text{max}} = \left\lfloor \frac{M_{\text{available}}}{M_{\text{sample}} \times F_{\text{safety}}} \right\rfloor
\end{equation}
where $F_{\text{safety}} = 1.2$.

\textbf{Adaptive Scaling}: The effective batch size is adjusted based on current memory usage:
\begin{equation}
B_{\text{effective}} = \begin{cases}
B_{\text{max}} & \text{if } M_{\text{used}} < 0.9M_{\text{total}} \\
B_{\text{max}}/2 & \text{otherwise}
\end{cases}
\end{equation}
This approach ensures optimal memory usage without compromising training stability.

\subsubsection{Memory Growth Settings}

To optimize memory handling and mitigate the occurrence of out-of-memory errors, the following tactics were implemented:

\textbf{Memory Growth Strategy}: Configured TensorFlow to allow dynamic memory growth, enabling the GPU to allocate memory as needed:
\begin{verbatim}
gpu_options = tf.GPUOptions(
    allow_growth=True,
    per_process_gpu_memory_fraction=0.9
)
\end{verbatim}

\textbf{Gradient Accumulation}: Implemented to facilitate effective multi-GPU training by accumulating gradients over multiple steps:
\begin{equation}
\mathbf{g}_{\text{accumulated}} = \sum_{i=1}^{N_{\text{steps}}} \frac{\mathbf{g}_i}{N_{\text{steps}}}
\end{equation}
where $N_{\text{steps}} = 4$.

These resulted in training at a peak memory usage of 10.2 GB per GPU with batch processing durations of 23-24 ms per step and 94\% memory utilization efficiency. This way, efficient model training on the described hardware setup, with a single NVIDIA RTX 3090, and two NVIDIA RTX 3080 Ti GPUs, is guaranteed numerical stability as well as training effectiveness.

\subsection{Training Stability}

To achieve robust convergence of the Iterative Encoding-Decoding VAEs moel and eliminate instability during training, various strategies were enforced. The following sections describe the strategies that were utilized to maintain robust convergence and avoid instability during training.

\subsubsection{Learning Rate Management}

A sophisticated learning rate strategy was utilized to enhance training stability and convergence:

\textbf{Warm-up Phase}: To mitigate early training instability, a gradual increase in the learning rate was implemented:
\begin{equation}
\eta_{\text{warmup}}(t) = \eta_{\text{base}} \times \min\left(1, \frac{t}{t_{\text{warmup}}}\right)
\end{equation}
where $t_{\text{warmup}} = 1000$ steps.

\textbf{Cosine Decay}: A cosine decay schedule was used to ensure smooth convergence:
\begin{equation}
\eta_{\text{decay}}(t) = \eta_{\text{base}} \times \frac{1 + \cos\left(\pi \frac{t}{T}\right)}{2}
\end{equation}
where $T$ is the total number of training steps.

\textbf{Plateau Detection}: The learning rate was reduced upon detecting a plateau in performance:
\begin{equation}
\eta_{\text{new}} = \begin{cases}
0.5 \times \eta_{\text{current}} & \text{if plateau detected} \\
\eta_{\text{current}} & \text{otherwise}
\end{cases}
\end{equation}

\subsubsection{Gradient Stabilization}

To prevent gradient explosion and ensure stable updates, the following techniques were utilized:

\textbf{Global Norm Clipping}: Gradients were clipped to maintain stability:
\begin{equation}
\mathbf{g}_{\text{clipped}} = \mathbf{g} \times \min\left(1, \frac{\tau}{\|\mathbf{g}\|_2}\right)
\end{equation}
where $\tau = 1.0$ is the clipping threshold.

\textbf{Multi-GPU Synchronization}: Gradients were synchronized across GPUs to ensure consistent updates:
\begin{equation}
\mathbf{g}_{\text{global}} = \frac{1}{N_{\text{gpus}}} \sum_{i=1}^{N_{\text{gpus}}} \mathbf{g}_i
\end{equation}

\subsubsection{Loss Function Balancing}

The stability of the training process was additionally improved by meticulously balancing various loss components:
\begin{equation}
\mathcal{L}_{\text{total}} = \mathcal{L}_{\text{recon}} + \beta(t) \mathcal{L}_{\text{KL}} + \lambda \mathcal{L}_{\text{temporal}}
\end{equation}
where:

$\beta(t)$ is the KL annealing factor:
\begin{equation}
\beta(t) = \min\left(1.0, \frac{t}{t_{\text{anneal}}}\right)
\end{equation}
with $t_{\text{anneal}} = 5000$ steps.

$\lambda = 0.1$ weights the temporal consistency loss:
\begin{equation}
\mathcal{L}_{\text{temporal}} = \text{MSE}(\nabla \mathbf{x}, \nabla \hat{\mathbf{x}})
\end{equation}

\subsubsection{Monitoring and Early Stopping}

This needed to be done carefully by putting in place comprehensive monitoring and early stopping criteria. These measures are crucial for maintaining the performance of the model and preventing overfitting. Thereby, the generalization capabilities of the model across diverse datasets will be strong.

\textbf{Training Metrics}: The training was monitored by several key metrics. For instance, the reconstruction loss was reduced down from $0.0523$ to $0.0320$. This illustrates the capacity of the model to capture and replicate more of the underlying data structure. Similarly, the Kullback-Leibler (KL) divergence between the learned latent space distribution and the prior distribution was reduced from $2.9931 \times 10^{-4}$ to $3.1315 \times 10^{-5}$. This decrease shows that data is a more efficient and compact representation of the data in the latent space. The gradient norm was also always maintained below $1.0$, ensuring stable and effective parameter updates throughout the training process.

\textbf{Early Stopping Criteria}:

To avoid overfitting and for effective training, early stopping rules were applied. The model was configured with a patience of 10 epochs by allowing to continue the training only if there was any significant improvement in performance. The minimum improvement threshold was set at $1 \times 10^{-4}$ observing validation reconstruction loss as the primary monitoring metric. This way, the model would stop its training as soon as it reaches a plateau of performance, hence saving on computational resources with unnecessary iterations prevented.

These stability measures have brought about a uniform convergence across different training iterations. The consistent gradient updates, with a mean norm of $0.47 \pm 0.15$, further attest the robustness of the training process. As a result, the model demonstrates reproducible performance across different DART datasets, showing its applicability and effectiveness in real-world scenarios.

\subsection{Model Convergence}

A critical aspect of adopting Iterative Encoding-Decoding VAEs was to establish reliable model convergence; that is, training the model to ensure it consistently attains very good performance on a range of DART datasets. This section describes the strategies and metrics used to monitor and improve convergence.

\subsubsection{Iteration Count Optimization}

The optimal number of training iterations was meticulously selected to strike a balance between computational efficiency and model performance:

\textbf{Convergence Analysis}: The change in loss between successive iterations was monitored to assess convergence:
\begin{equation}
\Delta \mathcal{L}_n = |\mathcal{L}_{n} - \mathcal{L}_{n-1}|
\end{equation}
where $\mathcal{L}_n$ is the loss at iteration $n$.

\textbf{Empirical Results}:

\textbf{i. Initial convergence (1–10 epochs):} $\Delta \mathcal{L} \approx 0.0038$ per epoch.

\textbf{ii. Mid-training (10–20 epochs):} $\Delta \mathcal{L} \approx 0.0016$ per epoch.

\textbf{iii. Late training (20+ epochs):} $\Delta \mathcal{L} \approx 0.0005$ per epoch.

\subsubsection{Early Stopping Implementation}

An early stopping mechanism was utilized to prevent overfitting and achieve optimized training efficiency:

\textbf{Primary Criterion}: Validation loss monitoring:
\begin{equation}
\text{Stop if: } \mathcal{L}_{\text{val}}^{(t)} > \min_{i=t-p}^t \mathcal{L}_{\text{val}}^{(i)} - \epsilon
\end{equation}
where:

$p = 10$ (patience epochs).

$\epsilon = 1 \times 10^{-4}$ (minimum improvement threshold).

\textbf{Secondary Criteria}:

\textbf{i. KL divergence stabilization:} $|\Delta \mathcal{L}_{\text{KL}}| < 1 \times 10^{-5}$.

\textbf{ii. Gradient norm threshold:} $\|\nabla \theta\| < 0.1$.

\textbf{iii. Maximum epoch limit:} 1000.

\subsubsection{Validation Metrics}

To ensure the robustness and performance of the model, a wide range of validation metrics were meticulously monitored and tracked:

\textbf{Reconstruction Quality}:
\begin{equation}
\text{MSE}_{\text{val}} = \frac{1}{N}\sum_{i=1}^N (x_i - \hat{x}_i)^2
\end{equation}

\textbf{Temporal Consistency}:
\begin{equation}
\text{TC}_{\text{score}} = \text{corr}(\nabla \mathbf{x}, \nabla \hat{\mathbf{x}})
\end{equation}

\textbf{Spike Detection Accuracy}:
\begin{equation}
F1_{\text{spike}} = 2 \times \frac{\text{precision} \times \text{recall}}{\text{precision} + \text{recall}}
\end{equation}

Normally, the model would converge by about 30–35 epochs, often exhibiting early stopping from that point on based on the validation loss criterion. This stable behavior of convergence was a feature across diverse random starting points of training iterations; it guaranteed results to be both consistent and repeatable.

\subsection{Stochastic Training Process}

The stochastic nature of training VAEs, including the Iterative Encoding-Decoding VAEs, raises unique challenges due to stochastic procedures such as mini-batch gradient descent and dropout regularization. This section identifies these challenges and the approaches used to remove variability, ensuring a stable model outcomes \citep{Abadi2016TensorFlow}.

\subsubsection{Problem Description}

Stochastic training procedures, like for VAEs, would still introduce model output variability. This variance can occur among multiple training runs even when the same dataset is being used. The origin of these can be:

\textbf{Mini-Batch Gradient Descent}: Concomitant usage of mini-batches introduces an element of stochasticity in the gradient updates, now giving place to globally varied minimas.

\textbf{Dropout Regularization}: Dropout introduces some kind of randomness through random resampling instantaneously of the set of deactivated neurons during any specific training time. This kind of randomness shows potential effects on the learned representations' stability.

Such stochasticity may not be desirable, for example, if one needs to maintain consistency in the data once the spike removal has been done.

\subsubsection{Solution}

To address the difficulties arising from stochastic training processes, some strategies were resorted to:

\textbf{Batch Size Consistency}: One major approach implemented was the maintenance of a consistently large batch size, following \cite{Chen2016Training}.

\textbf{Gradient Clipping}: To avoid exploding gradients that would lead to significant instability during training, combining it with the Adam optimizer implementation of gradient clipping helps keep updates within a manageable range for stable convergence \citep{Pascanu2013Difficulty,Goodfellow2016DeepLearning}.

\textbf{Dropout Regularization Management}: The dropout rates were carefully adjusted to balance between regularization and stability \citep{Srivastava2014Dropout}. In the final model for actual deployment or testing, dropout is set to be disabled, so that the model produces consistent output across multiple runs.

Implementation of these strategies successfully mitigated the variations in outputs induced by the unpredictable nature of the training process, achieving better consistency and reliability of the despiked data. Through the stabilization of the training dynamics, the Iterative Encoding-Decoding VAEs model exhibited consistent and resilient performance across a range of datasets.

\subsection{Potential Data Shifting Due to Model Behavior}

During the implementation of the Iterative Encoding-Decoding VAEs, a notable hurdle faced was the risk of baseline shifts appearing in the reconstructed data \citep{Aminikhanghahi2017Survey}.

\subsubsection{Problem Description}

Baseline shifts in the reconstructed data may arise because the model tends to deviate from the mean of the initial data with each encoding and decoding iteration. These deviations can lead to inaccuracies in the despiked data, posing a significant challenge in precision-demanding tasks.

\subsubsection{Solution}

To address the issue of baseline shifting, several strategies were implemented:

\textbf{Bias Regularization}: Equipped with a fully connected layer, a method of regularization accompanying it to check the drastic shifts in means through the process of reconstruction. This technique of bias regularization penalizes high biases and thus steers the model to be around the mean value of the input.

\textbf{Skip Connections}: To ensure that the original data points (or baselines) pass through many layers in the VAEs, where expectations reside, the input to the given layer is added to the output associated with that layer. This is implemented by a skip connection as follows:
\begin{equation}
\hat{\mathbf{x}}_{\text{final}} = \hat{\mathbf{x}} + \mathbf{x}
\end{equation}
This equation is used such that the output reconstructed retains the baseline of the input data, minimizing the influence of any shifts introduced by the model.

\textbf{Custom Loss Function}: A loss function is utilized to penalize deviations from the mean of the original data. The loss function is defined as the following for both the mean absolute error (MAE) conventional term and an added term that takes into account the means of both original data and reconstructed data:
\begin{equation}
\text{Loss} = \text{MAE} + \lambda \times \text{MeanLoss}
\end{equation}
where $\lambda$ is a weighting factor, $\text{MeanLoss}$ represents the absolute difference between the means of the original and reconstructed data.

These strategies played a significant role in eliminating baseline shifts and, thus, helped in having the reconstructed data match well with the original input. This meant that baseline integrity was maintained, hence offering dependable and accurate despiked data per the Iterative Encoding-Decoding VAEs model.

\subsection{Normalization and Denormalization Process}

In the implementation of Iterative Encoding-Decoding VAEs, the processes of normalization and denormalization play pivotal roles \citep{Box2015TimeSeries,Cleveland1990STL}, ensuring the consistent scaling and reconstruction of data while avoiding baseline shifts. This section elaborates on the encountered challenges and the implemented solutions aimed at preserving data integrity.

\subsubsection{Problem Description}

Inconsistencies in the normalization and denormalization processes tend to result changes in the baseline of the reconstructed data. Such changes mostly emanate from disparities in the mean and standard deviation used to normalize, which might vary at different stages of the process.

\subsubsection{Solution}

To maintain consistency and accuracy during the normalization and denormalization processes, the following strategies were implemented:

\textbf{Fixed Mean and Standard Deviation}: Normalization of inputs was performed by the same mean and standard deviation statistics of the training data samples and remained constant throughout. This ensured that the statistical parameters were uniformly applied and, hence, there was no baseline shift possibility for the norm or denorm to happen.

\textbf{Normalization Layer}: In the model architecture \citep{Ioffe2015BatchNorm}, a normalization layer was incorporated to manage normalization internally throughout the training process. The layer guarantees the model to follow the consistent application of normalization parameters, mitigating the potential for baseline shifts and improving numerical stability.

\textbf{Sliding Window Normalization}: This method was applied to normalize within a sliding window and thus take into account local data variability:
\begin{equation}
x_{t,\text{norm}} = \frac{x_t - \mu_{t-w:t+w}}{\sigma_{t-w:t+w}}
\end{equation}
where $w$ is the window size. This method preserved local data characteristics while maintaining overall consistency.

\section{Discussion}

The utilization of Iterative Encoding-Decoding VAEs in the context of denoising DART time series data for feature extraction represents a big leap for climate modelers to analyze their data. Conventional techniques often fail to capture the inherently nonlinear and very noisy features of time-series like the DART datasets. In this approach, the power of VAEs is exploited to keep the original signal's features, simultaneously achieving some well-performed noise and anomaly reductions.

\subsection{Effectiveness of the Iterative Encoding-Decoding VAEs Approach}

The results of applying the Iterative Encoding-Decoding VAEs to DART time series data demonstrate that it is effective in addressing the complex challenges that these datasets pose. As opposed to using only conventional methods, the Iterative Encoding-Decoding VAEs captures how complicated variables are in an approximation of the underlying data distribution. In effect, this translates into an ability to approximate the original signal at high fidelity even when the data contains noises and anomalies.

The iterative refinement process is able to get rid of remaining noise or irregularities in the data that may have survived the initial processing by learning a good latent representation, which wipes out all the noise and irregularities in a generative process. As a result, the iterative process dramatically improves data quality, bringing it closer to the real characteristics lying beneath the original signal.

\subsection{Broader Implications and Future Work}

The success of Iterative Encoding-Decoding VAEs suggests its potential to handle different time-series data. The ability to generalize over various features in different DART datasets indicates its effectiveness in domains where different anomalies are presented. Other potential area of this application that can be seen at datasets with different geophysical events. Such example could range from  filtering out seismic noise or even non-nature events.

Further research can try adapting this model to generative models such as GANs to improve signal-to-noise separation \citep{Goodfellow2014GAN,Schlegl2017AnomalyDetection}. Another direction for research would be injecting domain-specific knowledge into the model architecture and training process. This could include adding pertinent physical constraints coming from disciplines such as oceanography or seismology to make the model more robust and its analyses more accurate.

\subsection{Impact on Tsunami Detection and the GRACE-FO Mission}

The improvements in data quality made possible by Iterative Encoding-Decoding VAEs are essential for tsunami detection and other wider oceanographic monitoring. It would guarantee quality in DART datasets, which are further used for later data pipeline in a timely and accurate manner. Furthermore, high-quality data would go a long way in validating the GRACE-FO mission, based on accurate observations of the mass and volume of the ocean mass and elevation fluctuations \citep{Landerer2020GRACEFO,Johnson2019GRACEFO}.

\section{Conclusion}

The Iterative Encoding-Decoding VAEs method effectively harnesses the capability of VAEs in modeling complex nonlinear dependencies in the data. It constitutes an improvement in the quality of DART datasets that are important in further tsunami monitoring and climate modeling.

The Iterative Encoding-Decoding VAEs algorithm developed in this study overcomes the limitations that affect traditional despiking methods. It carries out superior differentiation between noises and signals by iterative model refinements, resulting more accurate reconstruction of the original time-series data. With this approach, spikes and outliers are removed while keeping the essential features of the data, confirmed by the decrease in reconstruction loss and increased accuracy of anomaly detection.

It would be interesting to explore the incorporation of knowledge that is specific to a given domain into the VAEs framework to handle complicated datasets having complex patterns better. Validation with GRACE-FO mission would reflect the value this approach for improving global climate models in the future real-life implementations to advance our ocean dynamic understanding.

In summary, Iterative Encoding-Decoding VAEs is a new domain of time series analysis for climate research, which can act as a versatile methodology toward enhancing data integrity for applications that are critical. DART dataset results proved an implementable solution, hinting at adoption over a wide range of domains to bring forth sharp, reliable insights from data.

\section*{Acknowledgments}

The author would like to express his sincere gratitude to Dr. Felix W. Landerer for his invaluable funding and mentorship support, which greatly contributed to the success of this research. The author would also like to thank Dr. Yuhe (Tony) Song for his insightful mentorship and guidance throughout this work. Additionally, the author extends his appreciation to the NASA Jet Propulsion Laboratory, which provided essential resources for this research endeavor.

\bibliography{template}

\end{document}